# Context and Humor: Understanding Amul advertisements of India

Radhika Mamidi

Language Technologies Research Centre (LTRC)
Kohli Center on Intelligent Systems (KCIS)
IIIT Hyderabad, Hyderabad
radhika.mamidi@iiit.ac.in

**Abstract.** Contextual knowledge is the most important element in understanding language. By contextual knowledge we mean both general knowledge and discourse knowledge i.e. knowledge of the situational context, background knowledge and the co-textual context [10]. In this paper, we will discuss the importance of contextual knowledge in understanding the humor present in the cartoon based Amul advertisements in India. In the process, we will analyze these advertisements and also see if humor is an effective tool for advertising and thereby, for marketing. These bilingual advertisements also expect the audience to have the appropriate linguistic knowledge which includes knowledge of English and Hindi vocabulary, morphology and syntax. Different techniques like punning, portmanteaus and parodies of popular proverbs, expressions, acronyms, famous dialogues, songs etc are employed to convey the message in a humorous way. The present study will concentrate on these linguistic cues and the required context for understanding wit and humor.

## 1 Introduction

Amul advertisements[1], hereby Amul ads, are unique and are a treat for a common man as well as a linguist. These Indian ads, appearing in billboard format, have been around for over 45 years. These advertisements are for the product butter. Amul Company has many other products ranging from milk to ice-creams, cheese to chocolates, and milk powder to beverages. By 2005, Amul entered the global market[2] as well. They also have many other commercials in different modes including videos. But the one for butter is the most popular and most consistent one. The billboards, placed at strategic locations in different cities of India, are changed on a weekly basis. The many blogs, articles and its fan groups on social networks like Facebook[3] reflect its popularity. Research work on these ads has been done by [11], [37], [39].

We have collected about 1250 ads for our study from Amul's website[4]. To analyze the ads, first we will discuss the important elements in the ad and then classify the ads based on different parameters including the pragmatic function. Then, we will look at the types of puns used which require contextual knowledge for understanding. As the target audience needs to be literate English-Hindi bilinguals and well-informed about the current events – politics, sports, films, social issues etc. and also have a good prior knowledge of popular songs, proverbs, sayings etc., wit may or may not be always the best marketing strategy.

---

[1] The brand name "Amul," from the Sanskrit "Amoolya," means "priceless". Formed in 1946, it is a dairy cooperative in India. It is managed jointly by the cooperative organization, Gujarat Co-operative Milk Marketing Federation Ltd. (GCMMF), and approximately 2.8 million milk producers in Gujarat, India.
[2] http://www.thehindubusinessline.com/todays-paper/tp-marketing/amul-seeks-a-slice-of-global-market/article2194983.ece
[3] https://www.facebook.com/amul.coop/
[4] http://www.amul.com/hits.html

## 2   Important elements in Amul ads

The main elements of Amul ads are the picture, the main text and the slogan. The picture is what catches the attention of the audience, what ignites the curiosity being always the key factor to read the message [6]. The contextual knowledge helps in forming the cohesive link between the textual message and the event depicted as a picture. The slogan usually refers to the event or people and links it with the product butter in a witty way.

### 2.1 The picture

Amul girl, the iconic figure with round eyes and blue hair [Fig. 7], transforms herself into different personalities or accompanies different personalities. The personalities are easy to recognize as they are often the one in the news in that week. The pictures may be a replica of the pictures found in newspapers or in posters as shown below in Fig.1 and Fig. 3.

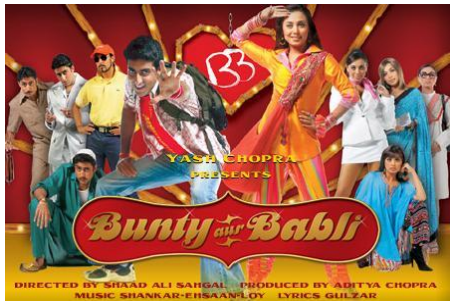
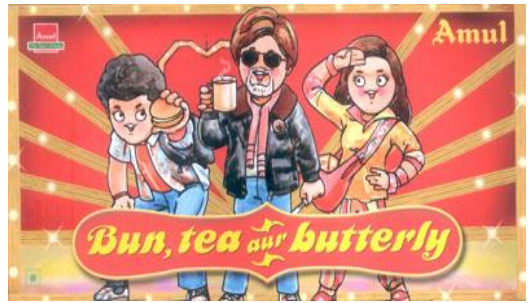

Fig. 1. Movie poster of 'Bunty aur Babli'          Fig. 2. Mimic of the movie poster 'Bunty aur Babli'

As can be seen, Fig. 2 is a replica of Fig. 1. If one is familiar with the poster, then one can make out that the figures in the ad are Abhishek Bachchan, Amitabh Bachchan and Rani Mukherjee.

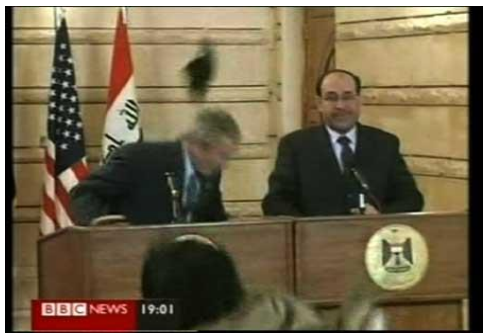
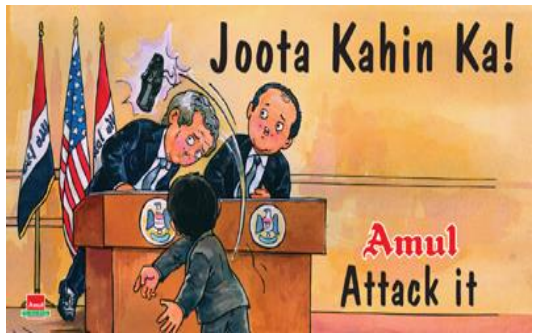

Fig. 3. Picture of Bush in newspapers          Fig. 4. Mimic of the shoe-hurling incident

Similarly, if one followed the news of an unhappy journalist hurling a shoe at President Bush, then one can figure out that that ad (Fig. 4) is about this event[5]. So, if one is abreast of current affairs, one can make out the target personality or event. Can you tell who the personalities in Fig. 5 and Fig. 6 are?

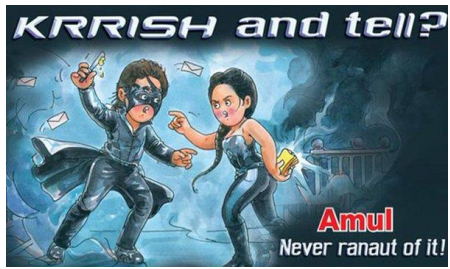 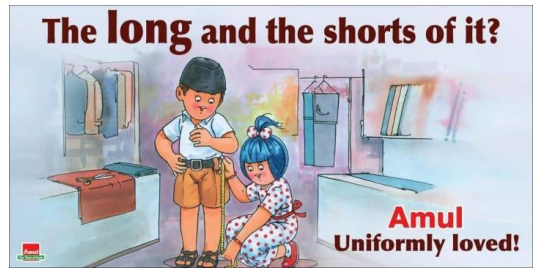

Fig. 5. Mimic of the movie poster 'Krrish 3'         Fig. 6. New uniform of RSS volunteers

### 2.2 The main text

Once the personalities in the images are resolved, one gets an idea about the event being referred to. Then, the wit in the text can be understood in a better way. In Fig. 4, the personality is George Bush and the main text "Joota kahin ka" (*shoe from somewhere*) is a pun on the word 'joota' which means shoe. There is an allusion to the commonly used phrase "jhooTha kahin ka" meaning '*liar from somewhere*' or *'What a liar!'*

Not always the text is understood so easily. For example, the next ad (Fig. 9) is based on the movie poster of 'Kaminey' (Fig. 8), but the text is understood only after watching the movie or knowing about the story in the movie. The main protagonists are twins and one of them has lisping problem and pronounces /s/ as /f/. So FUPER MAFKA! actually means 'super maska' ('maska' is *butter* in Hindi).

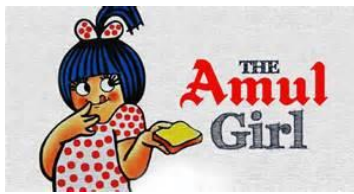    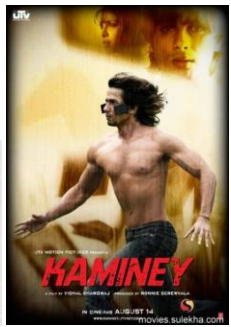    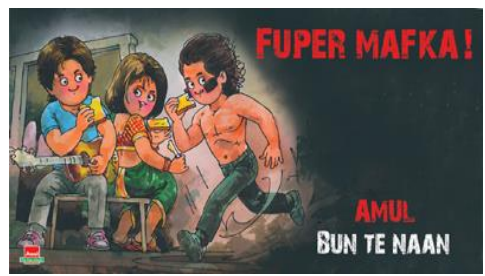

Fig. 7. The Amul mascot        Fig. 8. 'Kaminey' poster        Fig. 9. Mimic of 'Kaminey' movie poster

Similarly, in Fig. 5 if one identifies the personalities as Hrithik Roshan and Kangana Ranaut from the movie 'Krrish 3', it will be easy to recall the legal notices served by them to each other based on the 'kiss and tell' event[6]. And, of course, Fig. 6 has enough clues to show the personality as a RSS[7] (Rashtriya Swayamsevak Sangh) volunteer and the RSS' decision[8] to lenghthen the hem of the 90 year old dress code of khakhi shorts to trouser.

---

[5] https://en.wikipedia.org/wiki/Bush_shoeing_incident
[6] http://indiatoday.intoday.in/story/hrithik-kangana-fight-affair-details-timeline/1/812885.html
[7] https://en.wikipedia.org/wiki/Rashtriya_Swayamsevak_Sangh

### 2.3 The slogan/byeline

Slogans have an important role to play. By repetition, they become part of our memory and also everyday language [25]. The slogan for Amul ads is 'Utterly Butterly Delicious'. But it is not used always as seen in the ads above. Sometimes a parody of it is used as seen in the ad about 'Bunty aur Babli' (Fig. 2: *Bun, tea aur Butterly*), Facebook (Fig. 10: *utterly twitterly delicious*), other times it is substituted altogether with a different slogan as seen in the ad about Bt Brinjals[9] (Fig. 11 *Fully natural*). It is used in such a way that it also refers to the product butter. For example, in Fig. 4 *Attack it* refers to attacking Bush or attack the butter, *Fully natural* (Fig. 11) to the brinjals (eggplants) or to the butter.

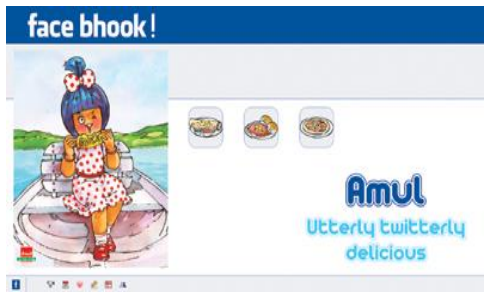
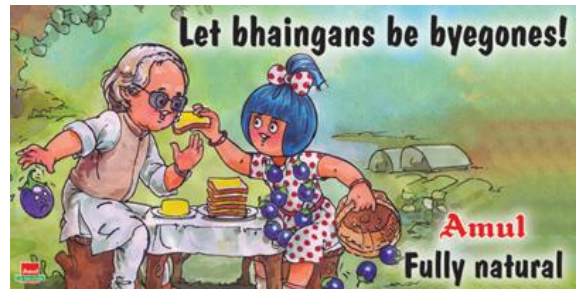

Fig. 10. Emergence of Social Media            Fig. 11. Bt Brinjals debate issue

### 2.4 The language used in the ads

According to Leech [16], the four principles of advertising texts are: Attention value, Readability, Memorability and Selling power. If the picture draws one's attention first, the language used for the main text and slogans play a key role in advertising. In a competitive world, the copywriters make sure their advertising texts are compact and all the elements are connected. The language of advertising throws light on a whole new kind of discourse [5], [14], [32].

An important element of Amul ads is its use of metaphorical switching or code switching between English and Hindi freely. The corpus helps us study the code used in the ad from over a period of time. If only monolingual English ads prevailed earlier, later Hindi made its way and there was code-mixing and code-switching, though the script remained Roman. Now, we have what we call as "Hinglish" which is equal use of English and Hindi elements blended as one single code. Hinglish is the language of the new generation[10].

Codeswitching is inevitable in a multilingual society [24]. Naturally, it is reflected in advertisements [11], [17], [18]. Hinglish or for that matter Tenglish (Telugu and English), Tamlish (Tamil and English) or Punglish (Punjabi and English) are English language blended with a regional language. The function of this code may be stated to be for proficiency in communication. This code is here to stay. The new generation wants to mark it as their own language; they want to decolonize English and give it an Indian flavour. Most of the international brands make use of Hinglish to relate closely with the Indian. We see that it is not just the words that are borrowed, but also the syntactic structures of Indian English. For example, the slogans for Coke "Life ho to aisi" (Life should be like this), Pepsi "Yeh dil maange more" (The heart wants more), McDonald's "What your bahana is?" (What's your excuse?) and Domino's Pizza "Hungry, kya?" (Are you hungry?). Also, most of the times, the variety of English used is Indian English which has its own quirks and specific phrases. In Amul ads, which reflect the current events, this change in code has been adopted by its copywriters as mentioned earlier. For example, the conjunction in *Bun, tea aur Butterly* (Fig. 2), in a compound formation *face bhook* (Fig. 10) where *bhook* means 'hunger' or a parody of the phrase *Let byegones be byegones* (Fig.11) in which the word 'byegone' is replaced by *bhaingans* meaning 'brinjals/eggplants'. [37] present a detailed study of the literary devices used in these codemixed ads emphasizing that "the use of Hinglish to juxtapose is to juxtapose two different cultures – the local and global, the traditional and modern, the indigenous and foreign."

---

[8] http://www.ndtv.com/india-news/rsss-new-khaki-pants-revealed-today-but-to-mixed-reviews-1451756
[9] http://www.esgindia.org/campaigns/press/say-no-bt-brinjal-say-no-release-genetic.html
[10] http://news.bbc.co.uk/2/hi/uk_news/magazine/6122072.stm

## 3  Is context based humor effective?

As seen from the previous sections, the popular Indian billboard advertisements which are topical in nature require the audience to be well-informed about the current events to identify the personality the moppet, known as the Amul girl/baby[11], is depicting or accompanying the personalities who have been targeted for that week.

In a competitive world, there is a pressure on copywriters to bring out ads that stand out from the rest. We come across many ads that are done so creatively that it leaves a good feeling in us. This positive feeling is what makes us receptive to the indirect persuasive function and makes us buy the product [3], [12]. The relevance-theoretic framework [28, 29] extended to study wittiness in advertising by [6] throws light on the aspect of the process of interpretation of witty advertising messages to be rewarding. The ads that are creative and innovative are remembered for a long time. If the message in the ad requires additional cognitive processing on the audience's part, it will increase its memorability [6], [21]. If the message in the ad is indirect and is intellectually satisfying and if the audience solve it, they feel happy for getting the witty message. This positive state of mind in return increases a positive attitude towards the product endorsed [7].

Studies have shown that the more attractive an advertisement is, the longer attention span it can command and lingers longer in one's memory [4]. The popularity of the witty Amul ads support the findings of the above studies. However, it contradicts with Dynel's [7] view that a witty advertisement hinders the interpreter's evaluation of the product as Amul butter is the most popular product in India mainly because its quality is maintained over the years since 1946[12]. It has withstood competition from international as well as indigenous brands like Nestle, Britannia and Mother Diary. It has also entered the world market and has established itself as a high quality product[13],[14]. It would be apt to reproduce the ad that came out when the gates were opened to the international market in early 1990s. Amul proudly proclaimed itself to be truly Indian [Fig. 12]. The message 'Be Indian. Bye Indian' makes an allusion to 'Be Indian, Buy Indian' slogan under colonial rule before India became independent in1947 revolting against foreign (British) goods made from Indian raw materials. The pun is on the word 'buy' making one rethink if globalization is good for the Indian economy. Amul ads try to sensitize us by using humor which makes us like it. In the next ad [Fig. 13], Amul makes us think if we are going forward or backward by banning Pakistani artists to work in Indian film industry. Here the pun is on the actor Fawad's name – Fawad/forward.

---

[11]https://en.wikipedia.org/wiki/Amul_girl
[12]http://www.india-reports.com/reports/Cheese3.aspx
[13]http://www.amuldairy.com/index.php/cd-programmes/quality-movement
[14] http://www.marketing91.com/swot-analysis-amul/

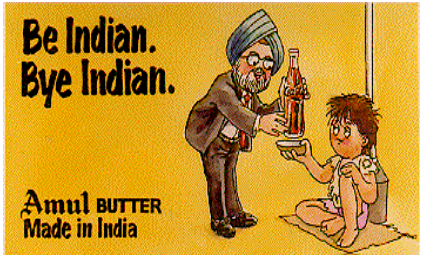 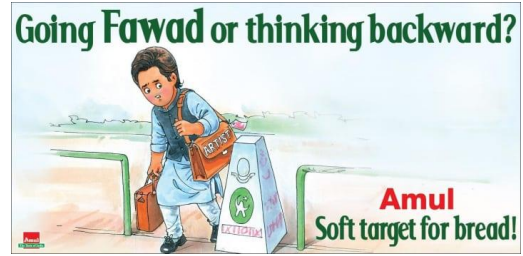

Fig. 12. Be Indian Buy Indian    Fig. 13. Pakistani artists banned

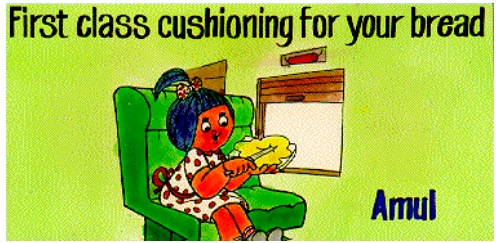 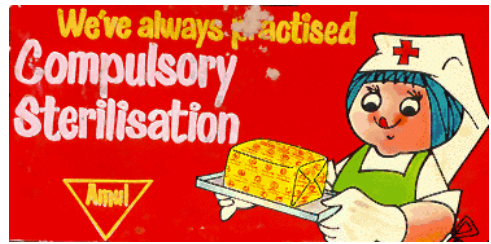

Fig. 14. Indian Railways    Fig. 15. Emergency time

In the remaining sections, we will come across many more examples depicting the rhetorical use of wit. By going through all the ads from the past 40 years and more one can get acquainted with the socio-political issues of modern India as seen in the two ads that appeared in the '70s [Figs. 14 and 15]. The ads, of course, will not be understood as we do not have the requisite contextual information that helps in the deeper cognitive processing of the ad to understand the wit and humor. For example, Fig. 14 refers to the introduction of cushioned chairs in the first class compartments by Indian Railways in 1979; and Fig. 15 refers to the compulsory sterilization introduced during Indira Gandhi's government in 1976. When the newspapers had also lost their voice during the Emergency period, Amul commending the act to reduce the population is laudable. But, the word 'compulsory' shows how wit is instrumental in bringing home the message as the family planning measures were supposed to be voluntary. In comparison, two of the latest ads [Figs. 16 and 17] based on a newly created word 'Covfefe' by President Trump and the most famous Indian movie Baahubali 'baaho se belly tak' (meaning *from arm to belly*) will need lesser prior knowledge. But, in 20 years, they may also become history and the need for the requisite knowledge to get the joke will be felt.

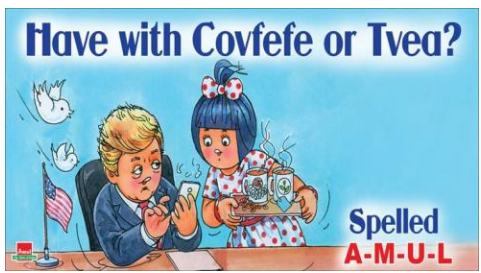 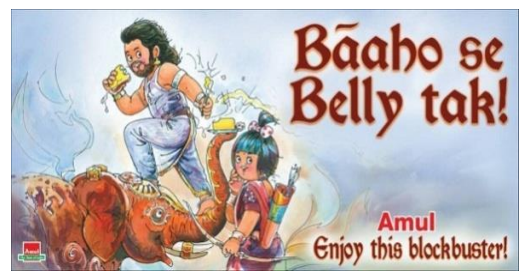

Fig. 16. President Trump's vocabulary    Fig. 17. Mimic of Baahubali 2 movie poster

## 4  Wit and Marketing

Humor in advertising has mixed statistics[15]. On one hand, it is shown that creativity and wit helps in the retention power of the product [34], [36], and on the other hand psychologists feel it colors the proper evaluation of the product, and the distraction may not help remember the product [35]. But, creativity in advertising was always appreciated. Some marketing analysts have given the credit for the popularity of Amul brand to the creativity factor found in the billboards. These billboards along with the brand itself have loyal customers from 2-3 generations. The company itself employs different strategies to promote all its products[16]. So given its strong position in the market, the copywriters have a field day playing with words as they don't have to worry about the persuasive function all the time as seen in the ads paying tribute [Figs. 18, 19] or condolence [Figs. 20, 21] which may not carry a witty remark or any reference to the product, but it definitely is remembered for the message. As the ads are topical in nature, one may study the changes in times with respect to technology, socio-economic reforms, political winds etc. by doing a diachronic study of these ads.

Over the years, the size of the word 'Amul' in the ads has also reduced. As the billboards occupy the same place, one knows that the ad is by Amul. The respect for a brand grows when it indulges in social message without marketing its product. The brand name is remembered for a longer period certainly.

According to Dynel [7], the two determinants of wittiness are *novelty* and *surprise*. These determinants are relevant to Amul ads as well, as seen earlier. The novelty is in the whole concept and the fact that the ads have a good following from over 40 years shows that it uses a unique technique to attract audience. Every ad has a surprise element that makes the audience chuckle.

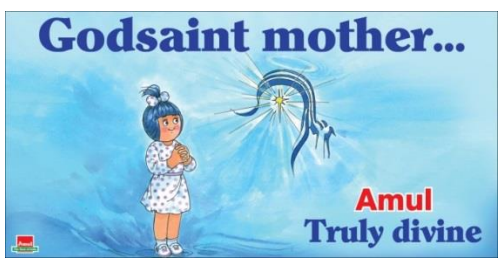

Fig. 18. Mother Theresa's canonization

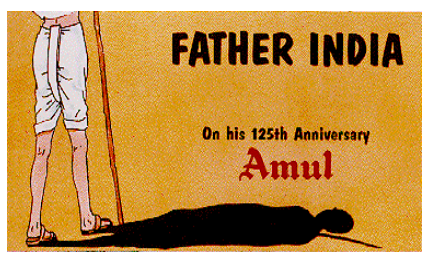

Fig. 19. Tribute to Gandhi

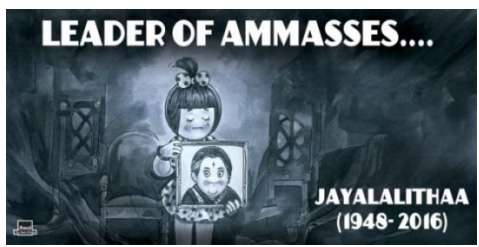

Fig. 20. RIP Jayalalitha

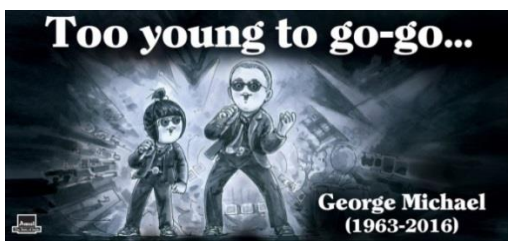

Fig. 21. RIP George Michael

---

[15] http://www.armi-marketing.com/library/LRE090121.pdf
[16] http://www.docstoc.com/docs/6464627/International-Marketing--Amul/

Understanding and perceiving humor is a cognitive process. Scholars from different fields especially from Linguistics, Philosophy and Psychology have been interested in theorizing it [9], [13], [15]. Theories of humor fall into any of the three kinds - release/relief theories, incongruity and superiority [19], [20], [23], [26].

Raskin [26] discusses several theories on humor including incongruity. Here we present some scholars views on incongruity. Monro [22, 23] calls "the importing into one situation what belongs to another" as incongruity. Mindess [20] adds: "in jokes… we are led along one line of thought and then booted out of it." Schopenhauer [27] puts forth a more consistent incongruity theory of humor. He suggests: "The course of laughter in every case is simply the sudden perception of the incongruity between a concept and the real objects which have been thought through it in some relation, and the laugh itself is just the expression of this incongruity". According to Sully [30], "the distinguishing intellectual element in humorous contemplation is a larger development of that power of grasping things together, and in their relation, which is at the root of all the higher perception of the laughable". In other words, in order to perceive incongruity there must be enough similarity between the events. The surprise element in a joke has been emphasized by incongruity theorists, too. The punchline of a joke presents this surprise element. It also provides the shift from one level of abstraction to another in a matter of seconds, and most noticeable it does seem incongruous with the main body of the joke. Bergson [2] proposes a special kind of incongruity theory which permeates all humor and is "something mechanical encrusted on the living". He believed that incongruity exists between the living and the automaton imposed on it.

By extending these theories viz. Relief, Incongruity and Superiority, to Amul ads, we see that *suspense* and *relief* are the key factors to the success of these advertisements. They cater to the curiosity of the audience. There are many followers that wait for the next advertisement to be out. The suspense is built in them. By placing the advertising hoardings at strategic traffic points, the humorous ads give the much needed relief to the stressed out Indian. This, in a way, supports the Relief theory.

Dynel [8] discusses in detail two approaches to humor interpretation - *Bisociation* and *Incongruity-Resolution (IR)* model. They deal with two unrelated stimuli that blend to produce humor. This is applicable to Amul ads where the theme in each ad is unrelated to the product ad. But most of the time the incongruity is resolved by the choice of words used that ultimately points to the product Amul butter as exemplified in section 2.3. In other words, the incongruity lies in the elements in the ad. For a person with the requisite contextual information, the incongruity is resolved by connecting the event to the product in the byeline. The slogans 'Uniformly loved' [Fig. 6], 'Fully natural' [Fig. 11], 'Truly divine' [Fig. 18], and 'Full of minerals' [Fig. 31] exemplify this.

Though the Superiority theory is not strong in this study, we see that these weekly ads demand a deeper cognitive analysis. As the ads require the target audience to be well informed about current events including national and international politics, sports, cinema, social issues etc, the author wonders if Amul ads is aimed at only a smaller target audience who are literate and follow news everyday either on television or newspapers. Given the language used is bilingual – Hindi and English, the audience number is further reduced in the multilingual country, India. They need to do a deeper cognitive processing to get the intended humor. This long process of cognitive processing also enhances the retention value. One of the reasons for the growing popularity of the ads in spite of such deeper cognitive processing is the smaller target audience (which given the population of India is not so small after all) that feel special. This feeling of 'being special', we feel, is no less than being Superior compared to those who do not get the joke.

## 5   Classifying Amul ads

Amul ads may be classified according to their themes or the pragmatic function or illocutionary force or the language used. Another way of classifying the ads is on the basis of the punning techniques employed in the main text. If we look at the themes, the ads can be classified as follows:

a. Sports: Cricket; Tennis; Badminton; Olympics
  b. Politics: Regional; National; International
  c. Films: Hollywood and Bollywood
  d. Current events/social issues: Swine flu; Bt brinjal; Narmada dam; scams and scandals

Based on their illocutionary force they may be classified as follows:
  a. Condemning: racial attacks; separation of states; wrong-doers; terrorist attacks
  b. Complimenting: Sportspersons; world leaders; city's spirit; new policies; festivals
  c. Creating Awareness campaign: to vote; to support a cause
  d. Mimicking/Acknowledging other contemporary products: Vodafone; Facebook; Ipod; Films
  e. Mourning: Death of celebrities – actors, singers, leaders

If we look at the code used in the ads, they may be classified as monolingual (only English) or bilingual (English and Hindi). The ads may be classified in a different way based on the punning element as will be seen in the sample analyses. Though we will focus on punning which is the trigger for generating humor, we tried to be representative of the different themes and pragmatic functions in the selection of our ads in this paper. It will be apt to mention a recent study on Amul ads [38] which classified the ads based on the three dominant colors used. They found that 1970s ads were very vibrant and colorful and the late 1990s had more of yellowish shades compared to the ads of 2000s.

## 6  Sample analysis

In this section, we will focus on the punning techniques employed in the ads. Punning is the most frequently employed technique in generating humor [1], [7], [31]. Puns are an important rhetorical device in advertisements [33]. To understand the humor generated using pun, a deeper cognitive processing is needed linking all the elements in our selected ads. An important external element is the contextual information which includes the general knowledge and linguistic knowledge as discussed earlier. The ads needs one to be up-to-date with the popular phrases, proverbs, movie dialogues etc. and a good knowledge of what is happening around. By studying the data we have, we see that there is a recurrent technique employed to generate humor.

We classified the ads based on puns into four types after studying our data consisting of 1250 (and more ads). We classified them based on sound/form (homophones and homographs), portmanteau or blending of two words, polysemous words and parody (of sayings, idioms, common phrases, acronyms, movie titles and songs etc). A sample analysis of these types is given below. We see that there is an overlapping of the classes. For example, the punning at sound level may also be using a portmanteau. In the Fig. 22, the ad is congratulating Pat Cash for a win over Ivan Lendl, who never won at Wimbledon. The pun is on the words C*ash* (polysemous – Cash and cash) and *Czech* (sounds as check). The byeline has a blended word *Lendlicious* (Lendl + Delicious) linking the event to the product. Similarly, the word *Devilopment* in Fig. 23 is by blending Devil and Development, condemning the error in the restructuring plan for Mumbai. This also reflects the pun at phonological level.

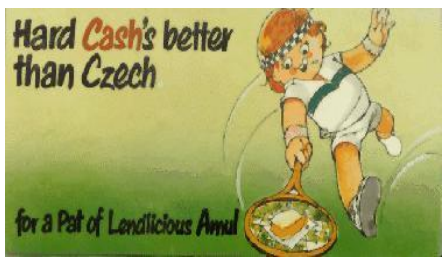 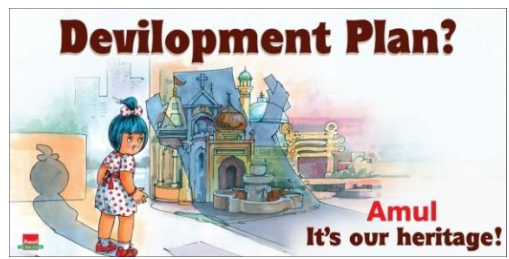

Fig. 22. Pat Cash beats Ivan Lendl at Wimbledon        Fig. 23. Restructuring Mumbai plan

### 6.1 Phoneme and grapheme level

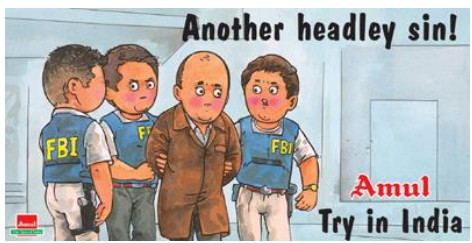
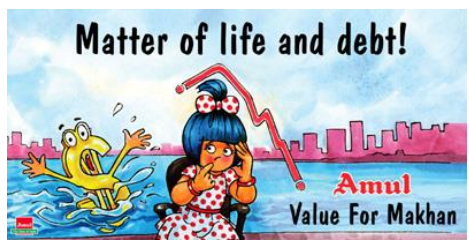

Fig. 24. Headley              Fig. 25. Financial crisis

In Fig. 24, the context is the proof found by FBI regarding David Headley's involvement in 26/11 terror attack in Mumbai - Dec.'09. The allusion is to the phrase *another deadly sin* where the pun is on the name *headley*. The byeline *Try in India* refers both to Headley's trial and persuading the audience to try the butter. The replacement of the word in a phrase by a rhyming word makes the pun to be understood easily. For example, in the ad (Fig. 25) about the financial crisis in the US regarding real estate, the word *debt* substitutes *death* in the phrase *matter of life and debt.* The phrase *value for makhan* in the byeline alludes to the product again (*makhan* = butter) along with the real estate (*makaan* = house). If these are puns at the phonological level, the ads below use puns at the grapheme level.

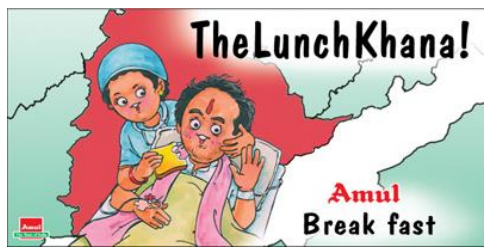
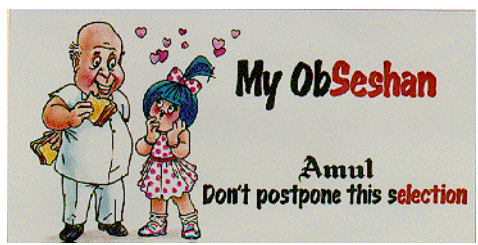

Fig. 26. Chandrasekhar's fast for a separate state       Fig. 27. Election Commissioner T.N. Seshan

In the above ad, the personalities are easy to identify. In Fig. 26, it is a regional party leader, Chandrasekhar, of Andhra Pradesh, going on hunger-strike demanding a separate *Telangana* state. The word Telangana is matched with *TheLunchKhana* (khana = to eat; The lunch khana = eating of the lunch). The pun is also in the byeline which is resolved by the *break* in the word *breakfast.* In the other ad (Fig. 27), there is a direct reference to the Election Commissioner of the late nineties, T. N. Seshan, who was liked by the common man for his policies. So *My Obsession* is a pun at the spelling level. The byeline refers to the general election as well as the selection of Amul butter. The ad uses different colors in the slogans to highlight the pun. In Fig. 18 is another example: God sent as *Godsaint* referring to Mother Theresa and in Fig. 5 *ranaut* is referring to (Kangana) Ranaut and run-out.

### 6.2 Portmanteau or Blending

In 2009, when H1N1 epidemic was creating confusion and panic among everyone, this ad [Fig. 28] aimed to promote awareness and prevention of the spread of the flu by the use of masks. The word *panicdemic* is created from panic and epidemic.

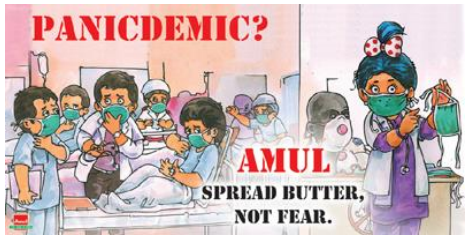 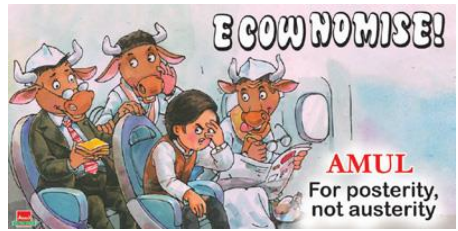

Fig. 28. The H1N1 epidemic fear          Fig. 29. Sashi Tharoor's comment on cattle class

The byeline *spread butter, not fear* puns on the word 'spread' linking the main text *Panicdemic* and the picture of a hospital to the product butter. If the context of the H1N1 virus and the panic associated with it is not available, then the audience may not get the full understanding of the ad and the new word. Similarly, the word *Ecownomise* in Fig. 29 refers to the minister, Sashi Tharoor's comment on Economy class as Cattle class and in Fig. 20 the word *Ammasses* is made from 'Amma' (meaning 'mother' - a name for late Chief Minister Jayalalitha used by her followers) and 'masses' who loved her.

### 6.3 Polysemous words

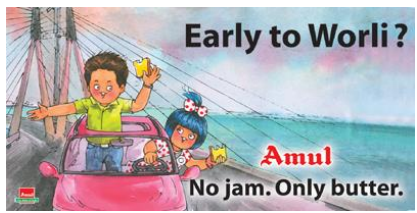 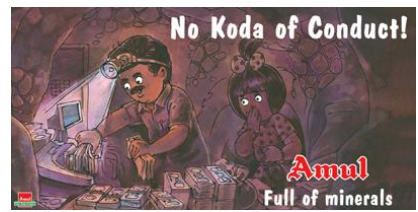

Fig. 30. Bombay's Worli Sea-link          Fig. 31. Madhu Koda – mining scam

The ad (Fig. 30) refers to the bridge across the sea linking Bandra to Worli. The bridge saved one from traffic jams. The required context includes that it refers to the bridge in Mumbai and the purpose behind constructing it. The byeline puns on the word *jam* – traffic jam and the edible jam. Another good example is the word *mineral* in *full of minerals* in the byeline of the ad (Fig. 31) which refers to the edible minerals in butter as well as to the mines – a reference to the mining scandal involving the chief minister of Jharkhand, Madhu Koda.

### 6.4 Parody

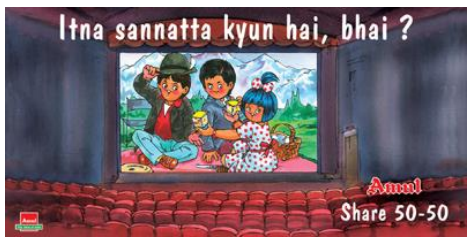 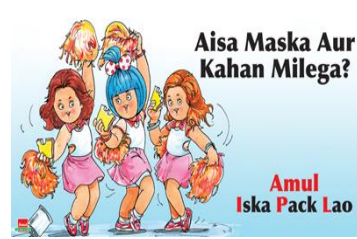

Fig. 32. Strike by multiplex theatres          Fig. 33. Cheer girls at IPL cricket matches

The main text 'No Koda of Conduct!' in Fig. 31 makes a reference to the English phrase *code of conduct* which is lacking in the chief minister Madhu Koda, who was involved in a mining scam. Another example of parody can be seen in Figs. 32 and 33. The main text in Fig. 32 translates to *Why is it*

*so quiet, dear?* The picture shows a theatre referring to the strike by multiplex theatre owners demanding a share of 50-50 from the film producers. With silence prevailing in the theatres, the text is apt. The wit lies in making the inter-textual reference to the dialogue from the film "Sholay" in which an old blind man utters these words to his neighbors who are silent as they are shocked looking at the dead body of the blind man's son. We have seen more examples of inter-textual references in the ads discussed above. A good knowledge of popular phrases, proverbs, movie songs, patriotic slogans etc is needed for understanding the inter-textuality. For example, Fig. 33 refers to an old song *Aisa mauka aur kahan milega* and in Fig. 21 to George Michael's *Wake me up before you go go*.

## 7  Conclusion

Humor is one of the best techniques used for marketing products. It creates a receptive attitude. The Amul ads of India are a perfect example of the use of humor in advertising. The ads form a good data to verify different theories of humor. Advertisers in India, including Amul, use bilingual techniques to relate to the modern Indian. The advertisements expect the audience to be up-to-date with the latest happening in the world. The relation between its popularity and the complex processing needs are indirectly linked. The incongruous elements in the ad are blended or resolved with the contextual knowledge. The humor thus generated makes the ad appealing and popular, thus providing a good evidence for humor as a good technique in marketing. The future work includes annotating all the ads based on the pragmatic function and humorous techniques as shown in the appendix.

**Appendix**

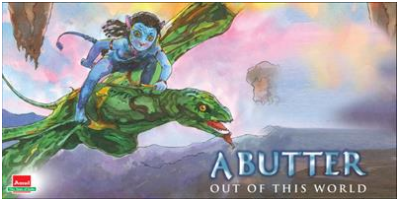

Event: Release of the movie 'Avatar '
Illocutionary force: Tribute
Language: Monolingual – English
Punning technique: Phonological punning on 'A butter' sounding as Avatar.
Resolution*: Out of this world* refers to the butter's quality/taste and the humanoids

Fig. 34. Mimic of Avatar movie poster

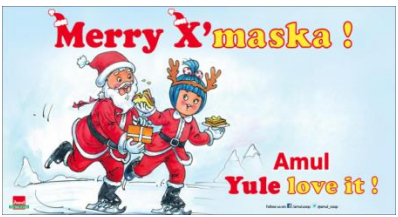

Event: Christmas
Illocutionary force: Greeting
Language: Bilingual
Punning technique: Compounding of two words X-mas and maska 'butter'.
Resolution:  *Yule* referring to Christmas and 'You'll' as in *You'll love it*. It referring to butter.

Fig. 35. Celebrating Christmas

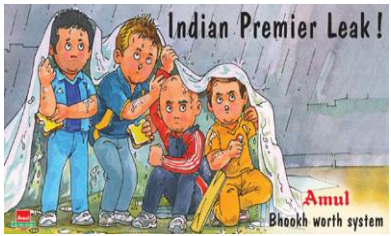

Event: IPL cricket match affected by rain
Illocutionary force: Empathy
Language: Bilingual
Punning technique: Parody of acronym IPL involving phonological punning on the word *League* of *Indian Premier League/Leak.*
Resolution: *Bhook (hunger) worth system* making an allusion to Duckworth-Lewis method, used to calculate the target score in cricket when the match is affected by weather or other circumstances.

Fig. 36. Washout of cricket match

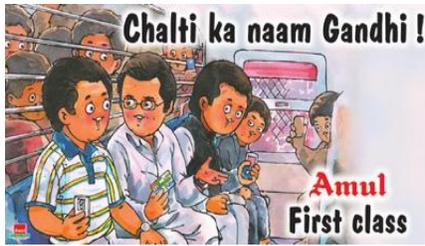

| Event: Rahul Gandhi travels by train<br>Illocutionary force: Commending<br>Language: Bilingual<br>Punning technique: Parody of an old movie tite 'Chalti ka naam gaadi'.<br>Resolution: *First class* refers to the classes found in train and the quality of the product. |

Fig. 37. Rahul Gandhi travels in local train

**Acknowledgements.** This work is supported by the Computational Humor Project no: LTRC-CPH-KCIS-78. I would also like to thank the anonymous reviewers for their comments that helped improve this paper.